# An Advancing Ensemble with Diversified Algorithms for Robot Arm Calibration

Zhibin Li, Shuai Li, *Senior Member, IEEE*, and Xin Luo, *Senior Member, IEEE*

*Abstract*—Recently, industrial robots plays a significant role in intelligent manufacturing. Hence, it is an urgent issue to ensure the robot with the high positioning precision. To address this hot issue, a novel calibration method based on an powerful ensemble with various algorithms is proposed. This paper has two ideas: a) developing eight calibration methods to identify the kinematic parameter errors; 2) establishing an effective ensemble to search calibrated kinematic parameters. Enough experimental results show that this ensemble can achieve: 1) higher calibration accuracy for the robot; 2) model diversity; 3) strong generalization ability.

*Index Terms*—Absolute Position Accuracy, Industrial Robots, Ensemble, Kinematic Parameters, Error Model.

## I. Introduction

INDUSTRIAL robots are the typical intelligent machines, their research are an important symbol to weigh the level of advanced manufacturing for a country [1-5]. They focus on the requirements of intelligent manufacturing, intelligent logistics, and intelligent life. However, due to the geometric error caused by the deformation of these robotic links and other factors, its absolute positioning accuracy is poor. Then it cannot be applied in high-precision manufacturing [6-10]. Thus, it is necessary to research the calibration of industrial robots [11-15].

In recent decades, numerous scholars have explored the application of optimization algorithms in robot calibration [15-20]. Gan *et al*. [1] propose a kinematic calibration method based on the drawstring displacement sensor, they adopt the least-squares method to identify the deviations of these kinematic parameters, which improves the calibration accuracy. Deng *et al*. [4] calibrate the industrial robots based on the MDH model. They propose a new calibration method combining LM algorithm and PF algorithm to calibrate the robot. Although these above methods can calibrate the robot, their calibration accuracy is poor. Can we adopt these algorithms to achieve an ensemble with higher calibration accuracy? To address this problem, this work researches eight calibration methods, such as EKF method [7], GA [13], SVM [29], LM algorithm [1], PF algorithm [4], EKF and PF algorithm [21], GA and LM algorithm, SVM and GA. Hence, we obtain eight advancing calibration models, which are further formed a strong ensemble [21-25].

This work contains the following main contributions:
a) Eight robot calibration methods with different algorithms are investigated, which can improve the calibration accuracy of robot;
b) This paper develops a powerful ensemble based on the eight different algorithms, which achieves the best performance in robot calibration.

For the rest of this paper, Section II introduces preliminaries. Section III presents the robot calibration methods based on eight diversified methods. Enough experiments and comparison are conducted in Section IV. Some conclusions are given in Section V.

## II. Preliminaries

### A. Problem Statement

In past decades, the commonly used kinematic model of robot is DH model, which can accurately and efficiently express the pose of robot end-effector [25-30]. Hence, we can obtain the transformation matrix of DH model [31-35].

$$^{i}T_{i+1} = Rot(z,\theta_i) \times Trans(z,d_i) \times Trans(x,a_i) \times Rot(x,\alpha_i)$$

$$= \begin{bmatrix} c\theta_i & -s\theta_i c\alpha_i & s\theta_i s\alpha_i & a_i c\theta_i \\ s\theta_i & c\theta_i c\alpha_i & -c\theta_i s\alpha_i & a_i s\theta_i \\ 0 & s\alpha_i & c\alpha_i & d_i \\ 0 & 0 & 0 & 1 \end{bmatrix} \tag{1}$$

where $c\theta_i = \cos\theta_i, s\theta_i = \sin\theta_i$, $^{i}T_{i+1}$ is link transformation matrix. *a*, *d*, *θ* and *α* are the link length, the link offset, the joint angle and

✧ Z. Li and X. Luo are with the School of Computer Science and Technology, Chongqing University of Posts and Telecommunications, Chongqing 400065, China (e-mail: LiZhibin111@outlook.com, luoxin21@gmail.com).
✧ S. Li and X, Luo are with Zienkiewicz Centre for Computational Engineering and Department of Mechanical Engineering, College of Engineering, Swansea University Bay Campus, Swansea SA1 8EN, UK (email: shuai.li@swansea.ac.uk, luoxin21@gmail.com).

the link twist angle, respectively.

The homogeneous transformation matrix from the base coordinate system to the end-effector coordinate is given as

$${}^0T_6 = {}^0T_1\,{}^1T_2\,{}^2T_3\,{}^3T_4\,{}^4T_5\,{}^5T_6 \tag{2}$$

Considering the differential principle, the equation (2) is transformed into

$$d\,{}^iT_{i+1} = \frac{\partial\,{}^iT_{i+1}}{\partial a_i}\Delta a_i + \frac{\partial\,{}^iT_{i+1}}{\partial d_i}\Delta d_i + \frac{\partial\,{}^iT_{i+1}}{\partial \theta_i}\Delta \theta_i + \frac{\partial\,{}^iT_{i+1}}{\partial \alpha_i}\Delta \alpha_i \tag{3}$$

According to equation (3), we can obtain the position error of robot end-effector.

$$dP = \begin{bmatrix} J_{11} & J_{12} & J_{13} & J_{14} \end{bmatrix} \begin{bmatrix} \Delta a \\ \Delta d \\ \Delta \theta \\ \Delta \alpha \end{bmatrix} = J \cdot X \tag{4}$$

where $dP$ is the position errors vector of robot end-effector, $X$ is the robot kinematic parameters errors vector, which is composed of $\Delta a$, $\Delta d$, $\Delta \theta$, $\Delta \alpha$. $J$ is the Jacobian matrix of the kinematic parameters [36-40].

In this work, the error between the nominal cable length $L_i'$ and the measured cable length $L_i$ is approximately equal to the position error of robot. Therefore, we can transform the least square problem of robot kinematic parameters optimization into

$$\min \|dP\|_2^2 = \min\left[\frac{1}{n}\sum_{i=1}^{n}(L_i - L_i')^2\right] \tag{5}$$

### III. THE IDENTIFICATION OF ROBOT KINEMATIC PARAMETERS BASED ON DIVERSIFIED ALGORITHMS

*A. EKF Algorithm*

The parameters identification equation of EKF algorithm is given by

$$X_{k|k-1} = X_{k-1|k-1},\ P_{k|k-1} = P_{k-1|k-1} + Q_{k-1} \tag{6}$$

where $X_k$, $P_k$ and $Q_k$ represent the deviations of the robot kinematic parameters, the covariance matrix and the covariance matrix of the system noise, respectively. We can obtain the position error $Z_k$.

$$Z_k = J_k X_k + E_k \tag{7}$$

$J_k$ is the Jacobian matrix. The Kalman gain calculation equation is written as

$$K_k = P_{k|k-1}J_k^T\left(J_k P_{k|k-1} J_k^T + R_k\right)^{-1} \tag{8}$$

where the covariance matrix of the measurement noise is $R_k$. Updating the estimation value of $X$, we have

$$X_{k|k} = X_{k|k-1} + K_k\left(Z_k - J_k X_{k|k-1}\right) \tag{9}$$

The optimal estimation of system covariance is

$$P_{k|k} = (I - K_k J_k) P_{k|k-1} \tag{10}$$

*B. LM Algorithm*

Least square method is a mathematical tool [1], which is widely used in numerous fields. Moreover, the LM algorithm is adopted to modify the ordinary least square method, we have

$$X = \left(J^T J + \lambda I\right)^{-1} \cdot J^T \cdot E \tag{11}$$

where $\lambda$ is the learning rate, $E = L_i - L_i'$.

*C. PF Algorithm*

The kinematic parameters are identified more accurately by PF algorithm [21]. Thus, we obtain the state-transition equation.

$$X_k = X_{k-1} + U_k,\ Y_k = H(X_N + X_k) - H(X_N) \tag{12}$$

$$H(X_N) = {}^0T_6 = {}^0T_1\,{}^1T_2\,{}^2T_3\,{}^3T_4\,{}^4T_5\,{}^5T_6 \Rightarrow H(X_N + X_k) = {}^0T_6 + \Delta\,{}^0T_6 \tag{13}$$

$Y_k$, $U_k$, $H$ represent the positioning error matrix of the robotic end-effector, the noise of this system and the forward kinematics operator, respectively.

According to the priori probability $p(X_0)$, we put particles $\{X_k^i\}$ into state space. $X_k^i$ represents the $i$th particle, whose value at the next time can be expressed as

$$X_k^i = X_{k-1}^i + U_k \tag{14}$$

By substituting (12) and (14), we obtain the position error for each particle.

$$Y_k^i = H(X_N + X_k^i) - H(X_N) \tag{15}$$

$Z_k^i$ represents the first three rows of the last column of $Y_k^i$, which is the positioning error of the robotic end-effector in $x$, $y$, $z$. Then we can obtain the error of nominal cable length, which is denoted as $L_k^i$. The weight of particles is calculated by the corresponding probability density, which is expressed as

$$\upsilon_k^i = \frac{1}{\sqrt{2\pi|R|}} \exp\left(-\frac{1}{2}\left[M_k - L_k^i\right]^T R^{-1} \left[M_k - L_k^i\right]\right) \tag{16}$$

The error of cable length measured by the cable actuated position encoder is $M_k$, $R$ is the covariance of noise. Based on (22), we achieve the normalization of particle weight.

$$\tilde{\upsilon}_k^i = \frac{\upsilon_k^i}{\sum_{j=1}^{N} \upsilon_k^i}, X_k = \sum_{i=1}^{N} \tilde{\upsilon}_k^i X_k^i \tag{17}$$

### D. SVM Algorithm

We adopt the SVM algorithm to address the issue of robot calibration [29], its loss function can be obtained as

$$\sigma = \min\left(\frac{1}{2}\lambda_1 \|w_2\|^2 + \frac{1}{2n}\sum_{i=1}^{n}(L_i - L_i')^2\right) \tag{18}$$

where $w_2$ and $\lambda_1$ are the regularization, learning rate, respectively.

To solve the regression problem based on SVM method, we introduce BP neural network. In addition, the input and output formulas of its hidden layer are given as

$$h_1 = w_1 K(\theta_i, \theta_j) + b_1, h_2 = g(h_1) = \frac{1}{1+e^{-h_1}}, L_i' = w_2 h_2 + b_2 \tag{19}$$

where, $h_1$, $h_2$ and $w_1$ are the input, output of layer and the weight for $K$, respectively. Here, $K(\theta_i, \theta_j) = \theta_i \times \theta_j$, $g(\bullet)$ is the activation function. Furthermore, $b_1$ and $b_2$ are the bias of the layer. Thus, we can obtain the weight update gradients of each layer network.

$$\frac{\partial \gamma}{\partial w_1} = -\sum_{i=1}^{n}\sum_{j=1}^{6}(L_i - L_i') w_2 h_2 (1 - h_2) K(\theta_i, \theta_j), \frac{\partial \gamma}{\partial w_2} = -\sum_{i=1}^{n}(L_i - L_i') h_2 + \lambda_1 w_2 \tag{20}$$

### E. Genetic Algorithm (GA)

GA is exploited to address the optimization problem of robot kinematic parameters, which has the advantages of simple algorithm, strong scalability and powerful global search ability.

### F. EKF and PF Algorithm (EPF)

EPF algorithm proposed by Jiang et al. [10] is a hybrid algorithm based on EKF algorithm and PF algorithm, which can successfully address the noise problem in the process of robot calibration [21].

### G. LM and GA Algorithm (LMGA)

To address the problem of multiple variables optimization, this paper proposes a method called "LMGA", which is based on GA and LM algorithm.

### H. SVM and GA Algorithm (SGA)

To identify the non-geometric error of the robot, we propose a new calibration method based on GA and SVM algorithm, which is called "SGA".

### I. Ensemble Method (EM)

In this work, we hope to obtain a strong ensemble based on the same principle for achieving a higher robot calibration accuracy. Then the boosting strategy for regression is employed to combine these basic models for developing a powerful ensemble.

## IV. EXPERIMENTS AND ANALYSIS

### A. General Settings

*1) Evaluation Metrics:* We employ root mean square error (RMSE), average error (Std) and maximum error (Max) as the evaluation metrics [41-46].

$$E_{Max} = \max\left\{\sqrt{(L_i - L_i')^2}\right\}, E_{Std} = \frac{1}{n}\sum_{i=1}^{n}\sqrt{(L_i - L_i')^2}, E_{RMSE} = \sqrt{\frac{1}{n}\sum_{i=1}^{n}(L_i - L')^2}, i = 1, 2, \cdots n \tag{21}$$

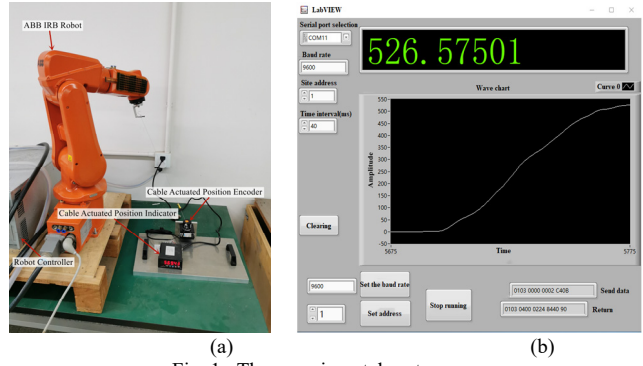
(a)                          (b)

Fig. 1. The experimental system.

TABLE I
COMPARED MODELS.

| Method | Description |
|---|---|
| ANN | The artificial neural network proposed in [7] that identifies the non-geometric errors of the robot. |
| BP | BP network proposed in [31], which is adopted to calculate controller angle errors in calibration. |
| UKF | The unscented Kalman filter proposed in [32], which is adopted to reckon the errors of the robotic kinematic parameters. |
| RBF | RBF neural network proposed in [33], which is constructed to estimate the positional errors of target positions. |

*2) Experimental Dataset:* In this work, 120 groups of samples are collected by the cable actuated position encoder, each of them contains the cable length and joint angle. Then, the 80%-20% training-test setting is adopted on this dataset.

*3) Experimental Process:* To achieve the calibration of the robot, we adopt the experimental platform shown in Fig. 1(a). Firstly, we choose the sampling points of the robot. Thereafter, the actual position is measured by the cable actuated position encoder. The interface of the data acquisition software is shown in Fig. 1(b). According to the measured dataset and the nominal kinematic parameters of the robot, the robot error is calculated by our algorithms. Lastly, we can obtain the actual kinematic parameters.

*B. Experimental Results of Eight Base Models*

Considering the experimental results of the eight base models, the important findings can be divided into the following points.

a) Commonly, LMGA algorithm has the highest calibration accuracy among all models. However, PF algorithm has the worst calibration accuracy. As shown in Table II, the RMSE, Std and Max of PF algorithm are 0.93, 0.80 and 1.81 respectively. Then, the RMSE, Std and Max of LMGA algorithm are 0.53, 0.44 and 1.28 respectively. Compared with PF algorithm, its accuracy gains are 43.01%, 45% and 29.28%, respectively.

b) The calibration accuracy of the combining model is higher than their base models. As recorded in Table II, the RMSE, Std and Max of EPF algorithm are 0.57, 0.46 and 1.52 respectively, which are 14.93%, 17.86% and 11.11% lower than 0.67, 0.56 and 1.71 by EKF algorithm, and 38.71%, 42.5% and 16.02% lower than 0.93, 0.80 and 1.81 by PF algorithm. Meanwhile, we obtain the similar experimental results in LMGA algorithm and SGA algorithm.

*C. Experimental Performance of the Ensemble*

In this part, we verify the experimental performance of the ensemble. Fig. 2 depicts the calibration accuracy of the ensemble via aggregating eight different algorithms. We can draw the following summaries from the experimental results.

a) The proposed ensemble has the obvious advantages in robot calibration accuracy. From Fig. 2, we see that the RMSE, Std and Max of ensemble are 0.49, 0.40 and 1.23 respectively. Compared with the RMSE, Std and Max at 0.93, 0.80 and 1.81 by PF algorithm, the improvement is 47.31%, 50% and 32.04% respectively.

b) Eight diversified calibration algorithms are beneficial to boost the calibration accuracy of ensemble. As shown in Fig. 2, the calibration accuracy is improved via aggregating eight different models into an ensemble sequentially.

*D. Comparison with Four Advanced Models*

In this part, the compared models are summarized in Table I, whose comparison results are described in Tables II. From them, we obtain the following important findings.

a) From Table II, the RMSE, STD and Max of ensemble are 0.49, 0.4 and 1.23 respectively, which are 18.33%, 20% and 6.82% lower than 0.60, 0.5 and 1.32 by BP algorithm. From Fig. 3 (a), the calibration accuracy of the ensemble with eight diversified algorithms is significantly higher than the above four advanced algorithms.

b) In the experiment, we collect 120 sample points for robot calibration. After calibration, we select the experimental results of some typical algorithms for comparison. Fig. 3 (b) and Fig. 3 (c) describe the robotic position error after calibration with 120 sample points by these algorithms. From their experimental results, the robot calibration accuracy of ensemble is the highest among all models, which verifies the feasibility of an ensemble.

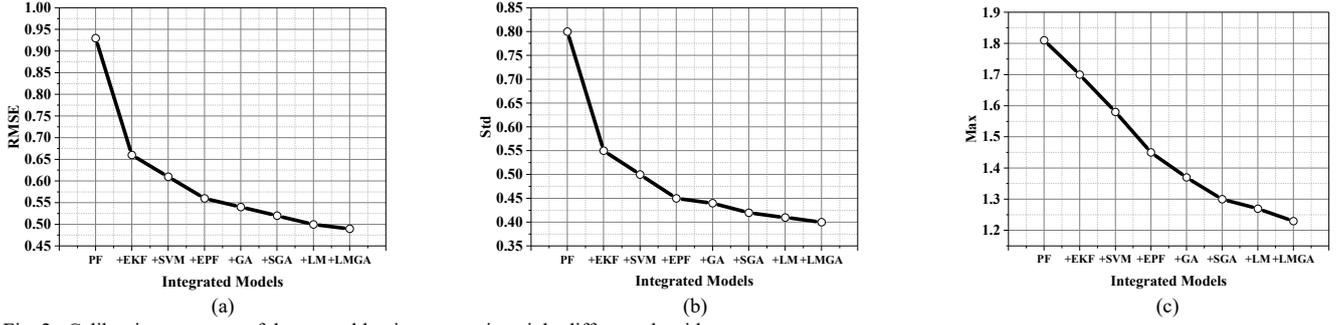
Fig. 2. Calibration accuracy of the ensemble via aggregating eight different algorithms.

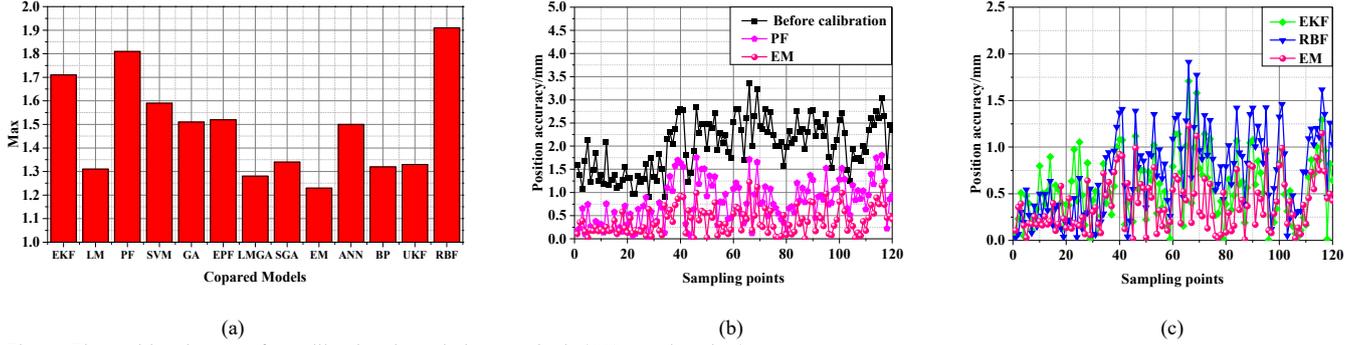
Fig. 3. The positional errors after calibration through these methods (120 sample points).

TABLE II
CALIBRATION ACCURACY OF COMPARED MODELS (120 SAMPLING POINTS).

| Item | RMSE(mm) | Std(mm) | Max(mm) |
|---|---|---|---|
| Before | 2.09 | 2.00 | 3.36 |
| EKF | 0.67 | 0.56 | 1.71 |
| LM | 0.55 | 0.46 | 1.31 |
| PF | 0.93 | 0.80 | 1.81 |
| SVM | 0.67 | 0.58 | 1.59 |
| GA | 0.61 | 0.48 | 1.51 |
| EPF | 0.57 | 0.46 | 1.52 |
| LMGA | 0.53 | 0.44 | 1.28 |
| SGA | 0.59 | 0.49 | 1.34 |
| EM | 0.49 | 0.40 | 1.23 |
| ANN | 0.64 | 0.55 | 1.50 |
| BP | 0.60 | 0.50 | 1.32 |
| UKF | 0.56 | 0.46 | 1.33 |
| RBF | 0.85 | 0.72 | 1.91 |

## V. CONCLUSIONS

In this paper, an efficient ensemble is developed. We first develop eight different calibration algorithms and analyze their calibration accuracy. Then, they are used as the base models to form an ensemble with boosting strategy. The experimental results show that the calibration accuracy of the ensemble is better than its base models and four advanced models.